\documentclass{article}
\usepackage{amsmath,epsfig}
\usepackage[preprint]{spconfa4}
\usepackage[dvipsnames]{xcolor}
\usepackage{cite}
\usepackage{marvosym}

\usepackage{times}
\usepackage{epsfig}
\usepackage{graphicx}
\usepackage{amssymb}
\usepackage{booktabs}
\usepackage{subcaption}
\usepackage{xcolor}
\usepackage{multirow}
\usepackage[pagebackref=true,breaklinks=true,colorlinks,bookmarks=false]{hyperref}

\let\OLDthebibliography\thebibliography
\renewcommand\thebibliography[1]{
  \OLDthebibliography{#1}
  \setlength{\parskip}{0pt}
  \setlength{\itemsep}{0pt plus 0.3ex}
}

\begin{document}\sloppy

\def\x{{\mathbf x}}
\def\L{{\cal L}}

\title{Adaptive Mean-Residue Loss for Robust Facial Age Estimation}
%
\name{Ziyuan Zhao$^{\dagger,\ddagger, \sharp}$, Peisheng Qian$^{\dagger}$, Yubo Hou$^{\dagger, \sharp}$  and Zeng Zeng$^{\dagger,\ddagger}$\textsuperscript{,\Letter}}
\address{
$^{\dagger}$Institute of Infocomm Research (I$^2$R), A*STAR, Singapore\\ $^{\ddagger}$Artificial Intelligence, Analytics And Informatics (AI$^3$), A*STAR, Singapore\\  $^{\sharp}$School of Computer Science and Engineering, Nanyang Technological University, Singapore \\
}

\maketitle

\begin{abstract}
Automated facial age estimation has diverse real-world applications in multimedia analysis,~\emph{e.g.}, video surveillance, and human-computer interaction. However, due to the randomness and ambiguity of the aging process, age assessment is challenging.  Most research work over the topic regards the task as one of age regression, classification, and ranking problems, and cannot well leverage age distribution in representing labels with age ambiguity. In this work, we propose a simple yet effective loss function for robust facial age estimation via distribution learning,~\emph{i.e.}, adaptive mean-residue loss, in which, the mean loss penalizes the difference between the estimated age distribution's mean and the ground-truth age, whereas the residue loss penalizes the entropy of age probability out of dynamic top-K in the distribution. Experimental results in the datasets FG-NET and CLAP2016 have validated the effectiveness of the proposed loss. Our code is available at~\url{https://github.com/jacobzhaoziyuan/AMR-Loss}.

\end{abstract}
\section{Introduction}
\label{sec:intro}

Automated facial age estimation has been widely applied to different multimedia application scenarios but remains a very challenging task. Human face aging is a complicated and random process affected by various internal factors, {\textit{e}.\textit{g}.}, genes, makeup, living environment. As a result, there are noticeable variances in facial appearance among different subjects with similar ages. Furthermore, the aging process of each subject is lasting, making it hard to perceive the variances in its facial appearance among neighbouring ages.

The research work for facial age estimation in the literature can be classified into three categories,~\emph{i.e.}, regression-based~\cite{AgeRegression2008,7780901}, classification-based~\cite{8784909}, and ranking-based~\cite{OrdinalRanker, AgeRanking}. But these methods fail to address the age distribution effectively, as there are no clear boundaries between adjacent ages, it is relatively easy for humans to guess the age with some confidence. For example, one person could guess a girl is around $25$ years old, or she may be in her mid-$20$s. In real life, the guessing process follows some certain probability distribution, which means that we could convert the mission of age estimation into a process of age distribution learning. In~\cite{CVPR2018Pan}, the authors assumed that the age distribution follows a Gaussian distribution with a particular mean age, $m_i$, and a standard deviation, $\sigma_i$, for $i$-th image, and proposed a novel mean-variance loss. The mean-variance loss attempts to penalize the variance between $m_i$ and the ground-truth age, $y_i$, by mean loss, and concentrate more on the classes around $m_i$ by variance loss. However, it is possible that the two loss functions suppress each other in part of the age distribution, achieving undesired solutions, and even worse, if $y_i$ happens to be out of the range $[m_i - \sigma_i,~ m_i + \sigma_i]$, variance loss penalizes the softmax loss that attempts to increase the probability of $y_i$. Furthermore, the age distributions of different subjects vary a lot, and it is not true that smaller $\sigma_i$ leads to a more accurate $m_i$.

\begin{figure}[tb]
    \centering
    \includegraphics[width = 7cm]{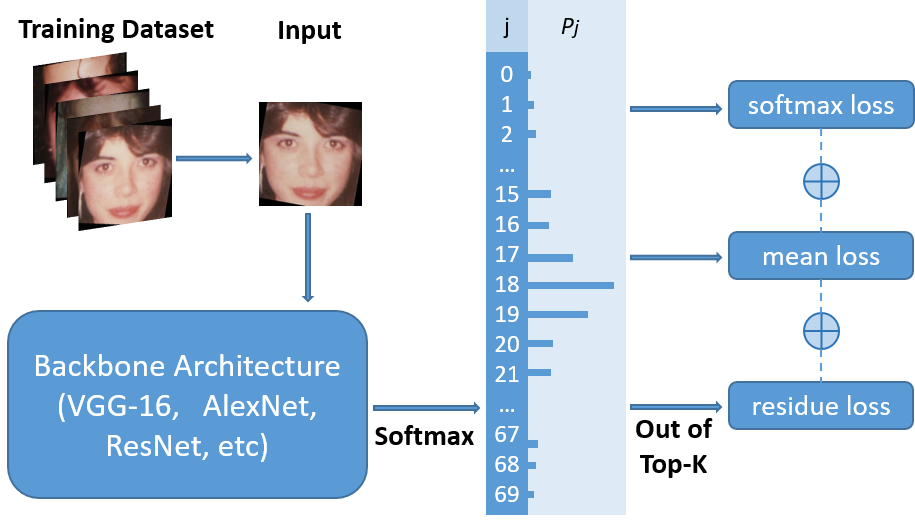}
    \caption{Overview of the proposed approach}
    \label{fig:approach}
\end{figure}

On the other hand, the top-K accuracy of deep learning models has achieved a very high level, \emph{e.g.}, the best \emph{top-5} accuracy on the ImageNet dataset is $98\%$~\cite{simonyan2014very}. In age estimation, we can assume that $y_i$ is included in top-K classes and the ages out of top-K have a very limited correlation with $y_i$, and then the sum of out-of-top-K probabilities can be treated as~\emph{residue}~\cite{TCYB}. However, how to determine K remains a problem. These motivate us to propose a hypothesis: \emph{If it is hard to extract deeper facial features, why not suppress uncorrelated features and dynamically penalize the residue to strengthen the correlation among the top-K classes indirectly?} 

In this work, we design an adaptive entropy-based residue loss, which can penalize the age probabilities out of dynamic top-K. By combining mean loss with residue loss, we proposed a simple, yet very efficient loss, adaptive mean-residue loss, for facial age estimation. The proposed mean-residue loss is simple to incorporate into other networks, such as ResNet. Experimental results are superior to the existing state-of-the-art benchmarks,~\emph{e.g.}, mean-variance loss.

\section{Related Work}

Early age estimation work was carried out in~\cite{AgeClassification94}, where ages were simply split into the following categories, {\textit{i}.\textit{e}.}, babies, young and senior individuals. Since then, more and more research interests have been attracted to age estimation from facial images~\cite{AgeRanking, AgeRegression2008}. Traditional methods for facial age estimation consist of two separate stages: feature extraction and one of age regression, classification, and ranking-based methods. In~\cite{FGNET}, an active appearance model (AAM) was proposed to use shape landmarks and textural features for age estimation. In~\cite{BIF2009}, age estimation is achieved by multi-direction and multi-scale Gabor filters with the feature pooling function were used for identification of biologically inspired features (BIF), but it still relied on feature representations crafted by hand, which is suboptimal to facial age estimation.

Deep learning based approaches have succeeded in many tasks,~\emph{e.g.}, object detection~\cite{FasterRCNN, wu2019interactive}, image classification~\cite{CVPR2015Res, qian2020multi}, biomedical signal processing~\cite{qu2020using,qu2020multi,ding2021learning}, and medical image analysis~\cite{yi2018enhance,zhao2021dsal,zhao2021mt}. They also have remarkably advanced facial age estimation recently. In~\cite{MultiscaleAge}, deep convolutional neural networks (DCNNs) were proposed to extract features from different regions on facial images and a square loss was utilized for age prediction. In~\cite{multitask2017}, a multi-task deep learning model was proposed for joint estimation based on numerous attributes, including shared feature extraction and attribute group feature extraction. To encode both the ordinal information between adjacent ages and their correlation, soft-ranking label encoding was proposed in~\cite{CVPR2019Tao}, which encourages deep learning models to learn more robust facial features for age estimation. 

On the other hand, label distribution learning (LDL) was proposed to distinguish label ambiguity, which is challenging.~\cite{LDL2016, TCYB}. In LDL, a label distribution can be assigned to an sample. Moreover, the correlation among values in the label space can be leveraged, from which a more robust estimation can be obtained. Gang~\emph{et al.} proposed several LDL based methods for age estimation and presented the robustness of them,~\emph{e.g.}, maximum-entropy modeling~\cite{LDL4Age}. It has been argued that a single facial image facilitates not only the estimation of a singular age, but is also informative for adjacent ages~\cite{LD4Facial}. In~\cite{CVPR2018Pan}, novel loss functions for model training were proposed and the mean-variance loss was proposed, in which the mean loss aims at minimizing the distance between the predicted age and ground-truth, while the variance loss attempts to lower the variance in the predicted age distribution. Consequently, the age distribution could be sharpened. Meanwhile, a sharper age distribution does not necessarily lead to better age estimation. Under particular circumstances, mean loss and variance loss contradict each other and prevent the accurate prediction to be achieved. In~\cite{TCYB}, residue loss was proposed to penalize residue errors of the long tails in the distribution for traffic prediction. We advanced the mean-residue loss in an adaptive manner for robust facial age estimation.  More details of loss analysis are discussed in Section~\ref{sec:gradient}. 

\section{Preliminary}
In an age estimation dataset, $y_i\in \{1,2, \ldots, L\}$ represents the corresponding age of $i$-th sample, $x_i$ stands for the facial feature, and $f(x_i)\in \mathbb{R}^{N\times M}$ represents the output from the layer, followed by a last fully connected (FC) layer. 
Let $z\in \mathbb{R}^{N\times L}$ be the output vector from the last FC layer, and $p\in \mathbb{R}^{N\times L}$ is the softmax probability defined in Equ.\ref{equ:softmax}:
\begin{equation}
\label{equ:softmax}
z = f(x_i)\cdot\theta^T; ~~~p_{i,j} = \frac{e^{z_{i,j}}}{\sum_{l=1}^Le^{z_{i,l}}},
\end{equation}
in which $x_i$ denotes the feature vector, $\theta\in \mathbb{R}^{L\times M}$ contains trainable parameters in the FC layer, $z_{i,j}$ is an element of $z$ in the $i$-th sample with age $j$. $p_{i,j}$ represents the probability that the age of sample $i$ is $j$.
Hence, the estimated or mean age $m_i$ of the sample $i$ is calculated in Equ.~\ref{equ:mean}:
\begin{equation}
\label{equ:mean}
E(age_i) = m_i = \sum_{j=1}^L j \cdot p_{i,j}.
\end{equation}

\section{Methodology}
\label{sec:methodology}
%
%

\begin{figure}[!thb]
    \centering
    \includegraphics[width = 0.45\textwidth]{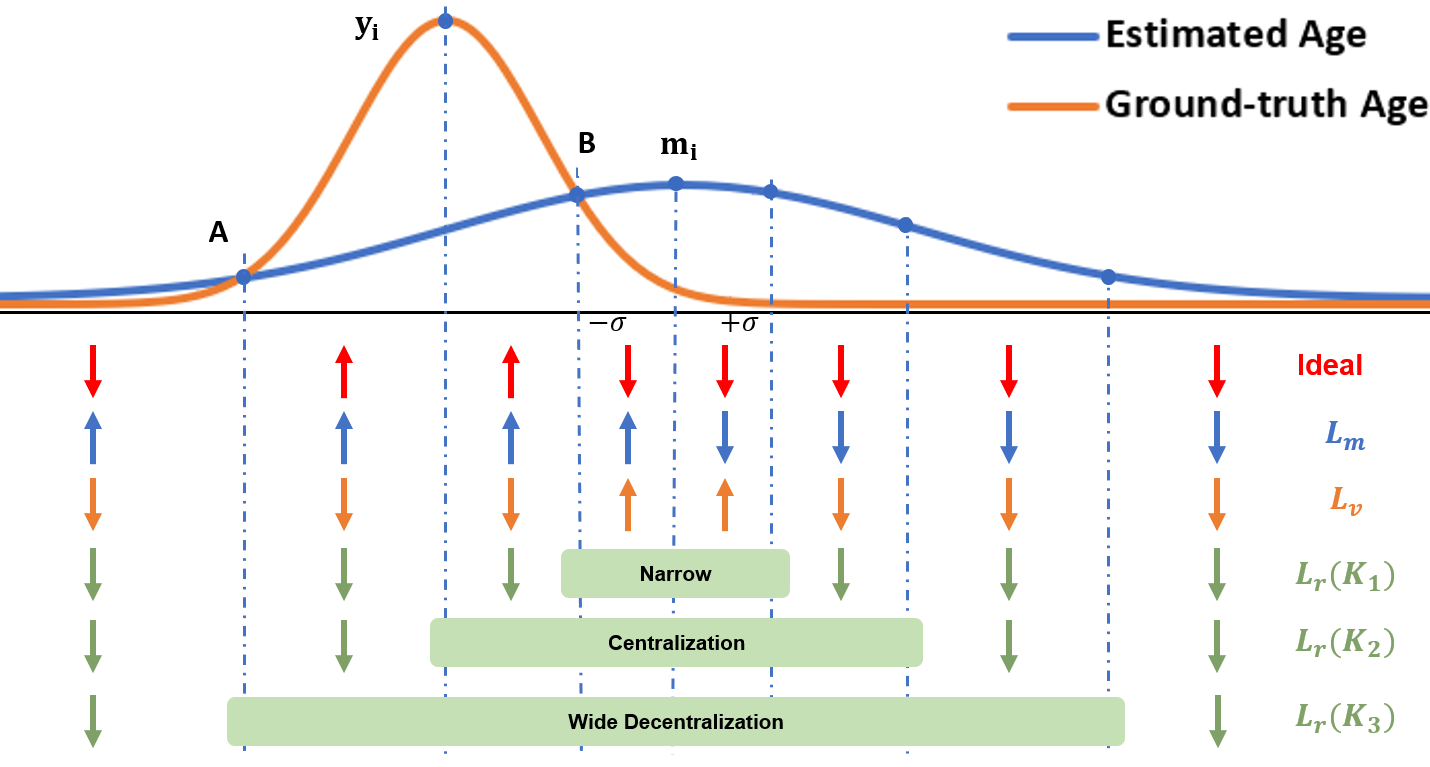}
    \caption{Gradient analysis on different components of the loss functions. (\textcolor{blue}{$L_m$}: the mean loss, \textcolor{orange}{$L_v$}: the variance loss, \textcolor{OliveGreen}{$L_r$}: the residue loss, where $K_1, K_2$ and $K_3$ denote three different situations.) $\uparrow$ and $\downarrow$ indicate the direction in which the loss function optimizes the distribution, while the ideal direction is indicated (\textcolor{red}{red}). For simplicity, $-\sigma$, $B$, and $K_1$ overlap; $y_i$ and $K_2$ overlap; $A$ and $K_3$ overlap.}
    \label{fig:gradient}
\end{figure}

As illustrated in~Fig.~\ref{fig:approach}, the proposed adaptive mean-residue loss can be embedded into a deep convolutional neural network (DCNN) for end-to-end learning.
As illustrated in Fig.~\ref{fig:gradient}, the proposed adaptive mean-residue loss penalizes (a) the difference between the mean of the estimated age distribution and the ground-truth age, and (b) residue errors in the two long tails of the age distribution.

\label{sec:gradient}
%

%
%
%
%

\subsection{Mean Loss}
The mean loss suppresses the variance between an estimated age distribution's mean and the true age.
According to Equ.~\ref{equ:mean}, we define the mean loss as
\begin{equation}
\label{equ:mean_loss}
L_m=\frac{1}{2N}\sum_{i=1}^N(m_i - y_i)^2 = \frac{1}{2N}\sum_{i=1}^N(\sum_{j=1}^L j \cdot p_{i,j} - y_i)^2,
\end{equation}
in which $N$, $m_i$, and $y_i$ are the training batch size, the estimated age, and and ground-truth age, respectively. Unlike the widely used softmax loss, the mean loss is presented in many regression problems. $L_2$ distance can be used to evaluate the variances between the mean of an estimated age distribution and the ground-truth age. Therefore, the proposed mean loss complements the softmax loss during training.

\subsection{Residue Loss} 
\label{residue_loss}
The residue loss penalizes the residue errors in the tails that exist in an estimated age distribution after the top-K pooling operation. 
%
%
The residue entropy is presented as follows, 
\begin{equation}
\label{equ:residue_loss}
   L_r = - \frac{1}{N}\sum_{i=1}^N ~\sum_{j=1, j~\notin top-K}^L p_{i,j} \cdot \log p_{i,j}. 
\end{equation}

The mean-variance loss~\cite{CVPR2018Pan} tries to suppress the probabilities within the range $m_i \in (j\in [1, m_i - \sigma]) \cup [m_i + \sigma, L])$, where $\sigma$ is the standard deviation of the age distribution in the $i$-th sample. However, inevitably, $y_i$ may probably fall inside this range. In this regard, $p_{i, y_i}$ is incorrectly suppressed, which could induce errors, while our residue loss $L_r$ guarantees that $y_i$ falls in the top-K classes and avoids such errors that occur with the mean-variance loss. To further optimize the residue loss, we propose a dynamic top-K pooling method. More specifically, let $K_i$ denote the number of top-K. Let $R_{y_i}$ stands for the ranking of $y_i$, for $i$-th image in a training batch. Subsequently, we can set $K_i$ to be adaptive as:
\begin{equation}
\label{equ:Kvalue}
K_i = max \{2, R_{y_i}\}.
\end{equation}

With the dynamic top-K pooling, $y_i$ is always included during the optimization process, and will not be incorrectly penalized by the residue loss.

\subsection{Adaptive Mean-Residue Loss} 
Combining with the softmax loss $L_s$, the adaptive mean-residue loss is showed in Equ.~\ref{equ:loss}:

%
\begin{equation}
\label{equ:loss}
\begin{aligned}
L&=L_s + \lambda_1 L_m + \lambda_2 L_r\\    
&=\frac{1}{N}\sum_{i=1}^N [-\log p_{i,y_i} + \frac{\lambda_1}{2}(m_i - y_i)^2 + \lambda_2 r_i],
\end{aligned}
\end{equation}
where $\lambda_1$ and $\lambda_2$ are hyperparameters that attempt to balance the influences of mean and residue sub-losses in the combined loss function. We use SGD~\cite{CVPR2015Res} to optimize parameters in the network. At inference, the estimated age of the $i$-th test image can be calculated based on Equ.~\ref{equ:mean}.

\subsection{Gradient Analysis}

\subsubsection{Comparison with Mean-Variance Loss}
The mean-variance loss and the mean-residue loss share the same mean loss function. The key difference exists in the variance loss and the residue loss, and their joint effect with the mean loss, as plotted in Fig.~\ref{fig:gradient}. In the mean-variance loss, the variance loss attempts to enhance the probabilities of classes within the range of $[m_i - \sigma, m_i + \sigma]$ while suppressing the probabilities of classes out of the range, \textbf{no matter $y_i$ is in the range or not}. When $y_i$ is out of the range, variance loss will decrease the probability of $y_i$, leading to a worse effect on the model performance. In the mean-residue loss, the residue loss attempts to suppress the probabilities of the classes out of top-K and help the network to focus more on the top-K classes indirectly. Adaptive top-K guarantees that $y_i$ is within top-K and residue loss function can help the network to increase the probability of classes within top-K as a whole. Hence, the performance of softmax loss can be enhanced.

\subsubsection{Situations of Different K}
The joint effect of the mean and residue loss is dependent on the choice of top-K in the residue loss. A proper selection of $K$ is the key to ensuring the correct optimization. The analysis of top-K is illustrated in Fig.~\ref{fig:gradient}, and different situations of top-K are summarized as follows: \textbf{(1) Over Centralization:} When top-K is too \textbf{narrow},~\emph{e.g.}, $K_1$ in Fig.~\ref{fig:gradient}, the ground-truth $y_i$ is \textbf{excluded} from the \textbf{over-centralized} top-K classes. The probability at $y_i$ is penalized by the residue loss in the wrong direction, and the residue loss conflicts with the mean loss at $y_i$, which is undesired. \textbf{(2) Centralization:} When top-K is \textbf{appropriate},~\emph{e.g.}, $K_2$ in Fig.~\ref{fig:gradient}, the ground-truth $y_i$ is \textbf{included} in the \textbf{centralized} top-K classes. The residue loss at $y_i$ is 0, and the joint effect of both mean and residue loss are the same as the ideal direction, which is desired. \textbf{(3) Decentralization:} When top-K is too \textbf{wide},~\emph{e.g.}, $K_3$ in Fig.~\ref{fig:gradient}, the ground-truth $y_i$ is \textbf{included} in the \textbf{decentralized} top-K classes. The joint effect of the loss functions is therefore following the ideal optimization direction. However, too many classes are covered by the top-K, in which the residue loss is not calculated. It lessens the effect of the residue loss and makes it more difficult to optimize the model.

Compared with a fixed $K$, the proposed dynamic top-K pooling can obtain a proper range for centralizing the age distribution. As the training proceeds, the distributions of the prediction converge towards the ground-truth, during which the optimal $K$ also decreases. Empirically, a fixed $K$ is either too small at the start of training, excluding $y_i$ from top-K pooling, or too large towards the end of the training, slowing down the optimization process. In contrast, the adaptive $K$, as shown in Equ.~\ref{equ:Kvalue}, always includes $y_i$ without over centralization or decentralization. Therefore it does not suffer from the drawbacks of a fixed $K$. From our theoretical analysis, we can anticipate that the adaptive mean-residue loss shall outperform the mean-variance loss in accuracy and convergence.

\section{Experiments}
\subsection{Datasets and Evaluation Metrics}
Extensive experiments have been carried out using the proposed loss on two popular facial image datasets,~\emph{i.e.}, FG-NET~\cite{FGNET_2014} and CLAP~2016~\cite{CLAP2016}. The ages of subjects lie between 0 to 69 years old.
We adopt the leave-one-person-out (LOPO) protocol in the experiments~\cite{CVPR2018Pan}. \textbf{CLAP2016}~\cite{CLAP2016} is a competition dataset released in $2016$ at the ChaLearn Looking at people challenges. There are $4113$ training subjects, $1500$ validation subjects and $1979$ test subjects. In CLAP~2016, An apparent mean age and standard deviation is labeled to each image.  For evaluation, we use the mean absolute error (MAE) between the ground-truth age $\mu_i$ and the prediction $y_i$ in FG-NET, while the $\varepsilon$-error is adopted from~\cite{CLAP2016} for CLAP2016.


%


\subsection{Experiment Settings} 
To reduce the influence of noises,~\emph{e.g.}, bodies, environments, all face images from different datasets were cropped with the cascaded classifier in OpenCV and resized into $256\times 256\times 3$.
%
%
We also employed data augmentation with rotation, flipping, color jittering, and affine transformations to reduce the overfitting.
We adopted VGG-16~\cite{simonyan2014very} and ResNet-$50$~\cite{CVPR2015Res} as our backbones for age estimation. 
The models are initialized with weights pre-trained using ImageNet~\cite{russakovsky2015imagenet}. 
The models are implemented using PyTorch. The initial learning rate and batch size are set to $0.001$ and $64$ respectively. 
The model is trained for $100$ epochs. Furthermore, for every $15$ epoch, the learning rate is reduced by a multiplication factor of $0.1$.

\subsection{Comparisons with Different Losses}
We compare the proposed mean-residue loss with the mean-variance loss proposed in~\cite{CVPR2018Pan}.
Both two losses have components softmax loss (i.e., $L_s$) and mean loss (i.e., $L_m$). 
Besides, we testify the effect of each component from both losses in an incremental manner. 
As shown in Table~\ref{table:fgnet}, we notice that the mean component always plays a core role in the prediction under both VGG-$16$ and ResNet-$50$. In comparison between variance loss and residue loss, the residue loss beats the variance loss in either case of the combination with the softmax loss or the combination with both the mean and softmax loss, which demonstrates the effectiveness of the proposed residue loss. Finally, our adaptive mean-residue loss outperforms all the other combinations, including mean-variance loss using VGG-$16$ and ResNet-$50$.

\begin{table}
\small
\centering
\setlength\tabcolsep{1.8pt}
\caption{Comparisons of different losses.}
\label{table:fgnet}
\scalebox{0.9}{
\begin{tabular}{c|c|c|c|c} 
\hline
\multirow{2}{*}{Method}      & \multicolumn{2}{c|}{FG-NET} & \multicolumn{2}{c}{CLAP2016}  \\ 
\cline{2-5}
                             & VGG-16        & ResNet-50         & VGG-16 & ResNet-50              \\ 
\hline
Softmax Loss                 & 7.19          & 6.99               & 0.4926 & 0.4756                      \\
Mean Loss + Softmax Loss     & 4.25          & 3.95              & 0.4687 & 0.4532                       \\
Variance Loss + Softmax Loss & 9.78          & 7.63             & 0.5516 & 0.5697                        \\
Mean-Variance Loss           & 4.10          & 3.95             & 0.4552 & 0.4018                       \\ 
\hline
Residue Loss + Softmax Loss  & 6.39          & 6.55             & 0.4921 &  0.4699                       \\
Adaptive Mean-Residue Loss   & \textbf{3.79} & \textbf{3.61}    &    \textbf{0.4511}    & \textbf{0.3882}     \\
\hline
\end{tabular}
}
\end{table}



\subsection{Influences of the Parameter $\lambda_1$ and $\lambda_2$}

\begin{figure}[!ht]
\centering
\includegraphics[width=0.4\textwidth]{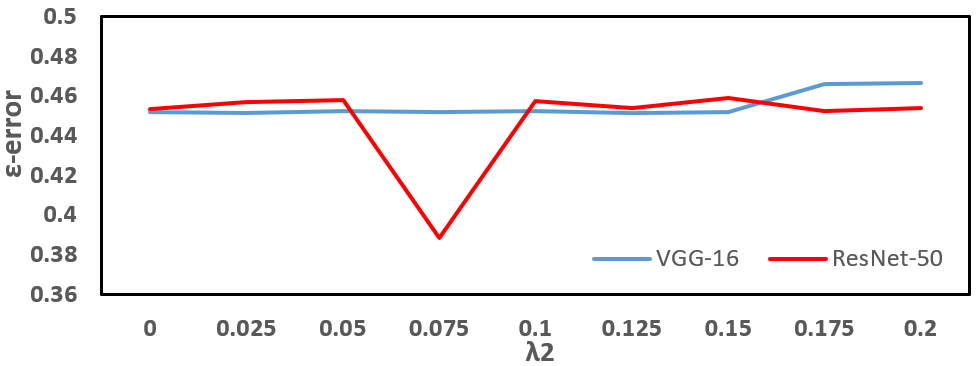}
\caption{The $\epsilon-$error w.r.t. $\lambda_1 =0.2$ and $\lambda_2$ from 0 to 0.2.
\label{fig:fig_lambda}
}
\end{figure}

Since hyperparameters $\lambda_1$ and $\lambda_2$ in Equ.~\ref{equ:loss} control the strengths of three components (softmax, mean, and residue) in the proposed loss during network training, we evaluate their influences of $\lambda_1$ and $\lambda_2$ on CLAP$2016$.
In the rest of this section, we aim to pick out the best portfolio for a pair of hyperparameters $\lambda_1$ and $\lambda_2$ in the proposed loss function. Following~\cite{CVPR2018Pan}, we set $\lambda_1$ to $0.2$ empirically and change $\lambda_2$ from $0$ to $0.2$ at interval of ever $0.025$. The $\epsilon-$error of the performance with different portfolios with different architectures,~\emph{i.e.,} VGG-16 and ResNet-50 are shown in Fig.~\ref{fig:fig_lambda}. We can see that VGG-16 is not sensitive to the changes of $\lambda_2$, while the best portfolio on CLAP~$2016$ is $\lambda_1 = 0.2$ and $\lambda_2 = 0.75$ for ResNet-$50$, which yields the lowest $\epsilon-$error.

\subsection{Influences of the Parameter K}

\begin{figure}[!thb]
\centering
\includegraphics[width=0.38\textwidth]{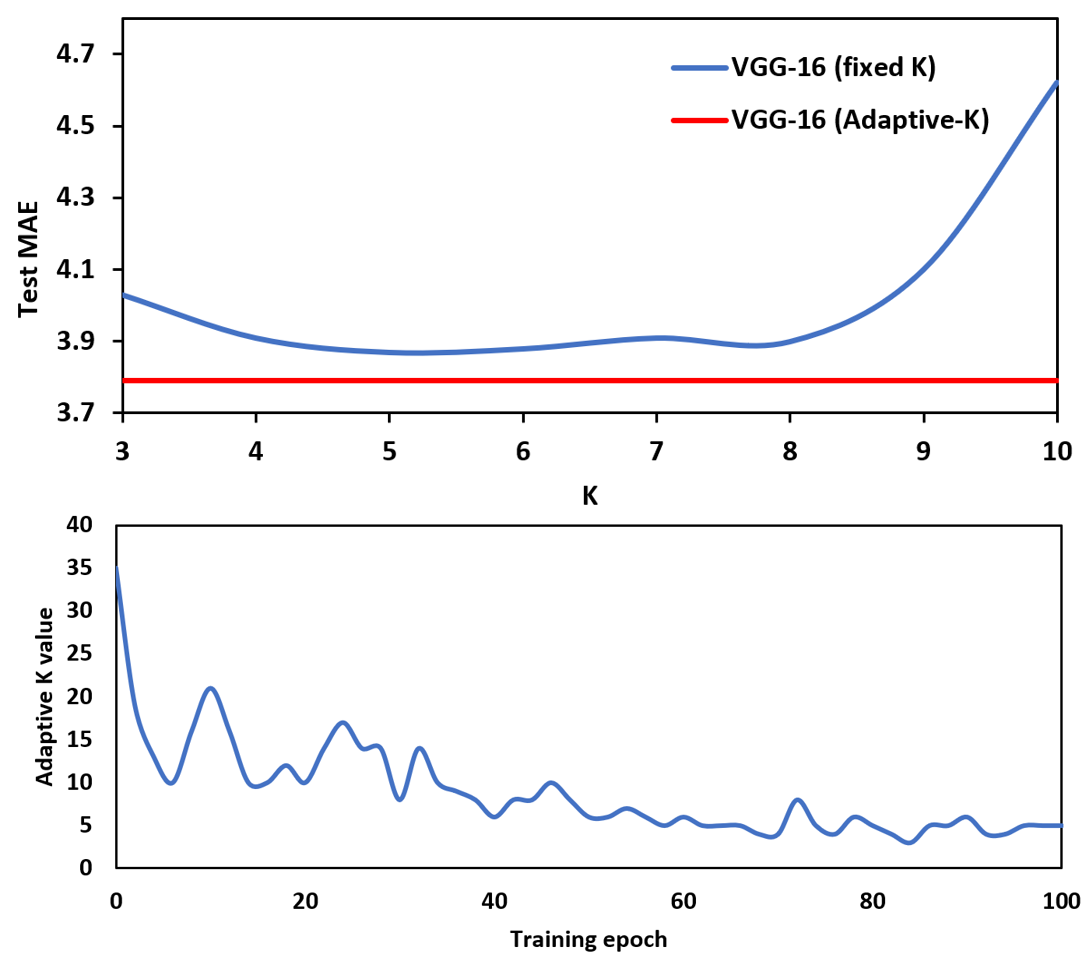}
\caption{Top: The MAE on FG-NET achieved by VGG-16 using different $K$ (\textcolor{blue}{blue line}) and Adaptive-K (\textcolor{red}{red line}). Bottom: The adaptive $K$ value during training.
\label{fig:combine}
}
\end{figure}

In Section~\ref{sec:gradient}, the influence of the choice of K on the model optimization is explained. Fig.~\ref{fig:combine} illustrates such impact of K values on the model performance. For a fixed $K$, it can be observed that the test MAE is lowest when $K=5$. Moreover, it is noted that the model with an adaptive K value consistently outperforms fixed K values. Fig.~\ref{fig:combine} describes the adaptive K values during training, which gradually converges to the best fixed K value of $5$, proving the capability of the algorithm to find the optimal $K$ during training. 

\begin{figure*}[!thb]
   \centering
    \includegraphics[width = 0.86\textwidth]{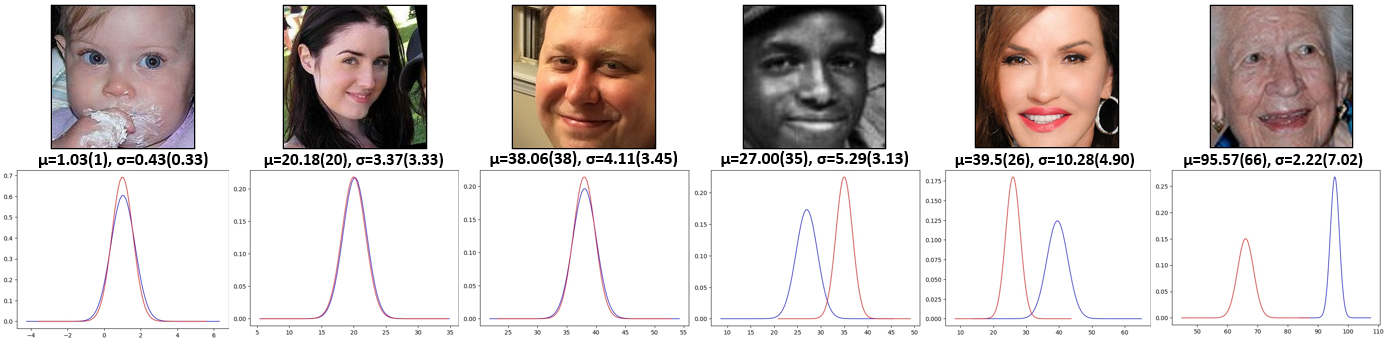}
    \caption{Examples with age distributions ($\mu$ = ground-truth age (estimated
age), $\sigma$ = ground-truth std (estimated std)) estimated by the proposed loss. The \textcolor{red}{red} and \textcolor{blue}{blue} curves are the \textcolor{red}{estimated} and \textcolor{blue}{ground-truth} distributions, respectively.} 
 \label{fig:visual}
\end{figure*}
\subsection{Comparisons with the State-of-the-art}
Experiments have been carried out for comparison between the proposed method and a number of state-of-the-art benchmarks on FG-NET and CLAP2016 respectively. As indicated in Table~\ref{tb:results_sota_fg}, the proposed mean-residue loss achieves the lowest MAE error among these approaches in FG-NET. Experimental results show that methods with distribution learning, such as mean-variance loss~\cite{CVPR2018Pan} can outperform ranking, regression, or classification based methods~\cite{chang2010ranking, OrdinalRanker, 7780901}. It is noted that, in DHAA~\cite{ijcai2019-492}, a hybrid structure with multiple branches was utilized to achieve the best performance among the rest approaches. We can explore the combination of DHAA with mean-residue loss in our future work. In Table~\ref{tb:results_sota_fg}, we directly quote the results of DeepAge and MIPAL\_SNU from~\cite{CLAP2016} as a complementary comparison. Our proposed loss outperforms mean-variance loss, which proves that residue loss with adaptive-K pooling is helpful to concentrate more on top-K ages indirectly. In Fig.~\ref{fig:visual}, we present some examples (3 good cases and 3 poor cases) predicted by the proposed method on CLAP~$2016$. The proposed approach performs well for different age groups. But when the images have poor qualities,~\emph{e.g.}, bad illumination, blurring, the age estimation accuracy decreases dramatically. In addition, good makeup and extreme values would influence the results.

\begin{table}[!thb]
\small
\centering
\caption{Comparisons of different methods}
\label{tb:results_sota_fg}

\setlength\tabcolsep{10pt}
\scalebox{0.9}{

\begin{tabular}{|l|c|c|}
\hline
\multicolumn{3}{|c|}{FG-NET dataset}                                \\ 
\hline
Method                   & MAE            & Protocol \\ \hline
RED-SVM~\cite{chang2010ranking}                   & 5.24          & LOPO    \\
OHRank~\cite{OrdinalRanker}                   & 4.48         & LOPO     \\
DEX~\cite{7780901}                      & 4.63           & LOPO    \\
Mean-Variance Loss~\cite{CVPR2018Pan}       & 3.95           & LOPO     \\
DRFs~\cite{8578343}                     & 3.85           & LOPO   \\
DHAA~\cite{ijcai2019-492}                     & 3.72          & LOPO      \\ \hline
Adaptive Mean-Residue Loss & \textbf{3.61}              & LOPO     \\ \hline
\multicolumn{3}{|c|}{CLAP2016 dataset}                                \\ \hline
Method                   & $\varepsilon$-error  & Single Model? \\ \hline
DeepAge~\cite{CLAP2016}                 & 0.4573 & YES            \\
MIPAL\_SNU~\cite{CLAP2016}      & 0.4565 & NO           \\
Mean-Variance Loss~\cite{CVPR2018Pan}               & 0.4018  & YES           \\ \hline
Adaptive Mean-Residue Loss & \textbf{0.3882}      & YES           \\ \hline
\end{tabular}
}

\end{table}


\section{Conclusion}
%

In this paper, we propose a simple, yet very efficient mean-residue loss for robust facial age estimation. We verify the superiority of our proposed method over state-of-the-art benchmarks through theoretical analysis and experiments. In the future, we would extend our method to other domains for continuous value estimation, such as survival year estimation in healthcare and sales prediction in e-business.

\section{Acknowledgement}

We are grateful for the help and support of Xiaohao Lin, Xi Fu and Ce Ju. The work is supported by Institute for Infocomm Research (I$^2$R) and Artificial Intelligence, Analytics and Informatics (AI$^3$), A*STAR, Singapore.

\small
\bibliographystyle{IEEEbib}
\bibliography{icme2022template}

\end{document}